\documentclass[10pt,twocolumn,letterpaper]{article}

\usepackage[pagenumbers]{cvpr} 
\usepackage[pagebackref,breaklinks,colorlinks,citecolor=cvprblue]{hyperref}
\usepackage[dvipsnames]{xcolor}
\usepackage{graphicx}
\usepackage{times}
\usepackage{epsfig}
\usepackage{amssymb}
\usepackage{booktabs}
\usepackage{nicefrac} 
\usepackage{multirow}
\usepackage{bm}
\usepackage{bbm}
\usepackage{kotex}
\usepackage{comment}
\usepackage{dsfont}
\usepackage[dvipsnames]{xcolor}
\usepackage{soul}
\usepackage{algorithmic}
\usepackage{setspace}
\usepackage[ruled,vlined]{algorithm2e}
\usepackage{bm}
\usepackage{bbm}
\usepackage{color}
\usepackage{soul}
\usepackage{float}
\usepackage{enumitem}
\newcommand{\sysname}{eBTS}
\definecolor{cvprblue}{rgb}{0.21,0.49,0.74}

\title{Class-Wise Buffer Management for Incremental Object Detection: \\ An Effective Buffer Training Strategy}

\author{%
    Junsu~Kim\textsuperscript{1}%
    \quad Sumin~Hong\textsuperscript{2}%
    \quad Chanwoo~Kim\textsuperscript{1}%
    \quad Jihyeon~Kim\textsuperscript{1}%
    \quad Yihalem~Yimolal~Tiruneh\textsuperscript{1}%
    \\ Jeongwan~On\textsuperscript{4}%
    \quad Jihyun~Song\textsuperscript{3}%
    \quad Sunhwa~Choi\textsuperscript{3}%
    \quad Seungryul~Baek\textsuperscript{1} \\\\
    \textsuperscript{1}UNIST, South~Korea 
    \quad \textsuperscript{2}SeoulTech, South~Korea \\
    \quad \textsuperscript{3}LG Electronics, South~Korea
    \quad \textsuperscript{4}Chonnam National University, South~Korea\\
}

\begin{document}
\maketitle
\begin{abstract}
    Class incremental learning aims to solve a problem that arises when continuously adding unseen class instances to an existing model This approach has been extensively studied in the context of image classification; however its applicability to object detection is not well established yet. Existing frameworks using replay methods mainly collect replay data without considering the model being trained and tend to rely on randomness or the number of labels of each sample. Also, despite the effectiveness of the replay, it was not yet optimized for the object detection task. In this paper, we introduce an effective buffer training strategy (\sysname{}) that creates the optimized replay buffer on object detection. Our approach incorporates guarantee minimum and hierarchical sampling to establish the buffer customized to the trained model. 
    Furthermore, we use the circular experience replay training to optimally utilize the accumulated buffer data. Experiments on the MS COCO dataset demonstrate that our \sysname{} achieves state-of-the-art performance compared to the existing replay schemes.
\end{abstract}    
\section{Introduction}
\label{sec:intro}
\vspace{-0.5em}
Traditional machine learning models tend to forget previously learned patterns when trained on new datasets, a phenomenon called ``catastrophic forgetting"~\cite{robins1995catastrophic}. This poses challenges for models operating in dynamic environments. However, unlike machines, humans can learn new concepts without entirely forgetting pre-existing knowledge. Building on this insight, incremental learning aims to address this issue by training models to assimilate new concepts progressively without retraining on the entire past dataset, effectively preserving knowledge from prior tasks while integrating new task.
\begin{figure}[!t]
\setstretch{0.9}
\centering{
\includegraphics[width=1\linewidth]{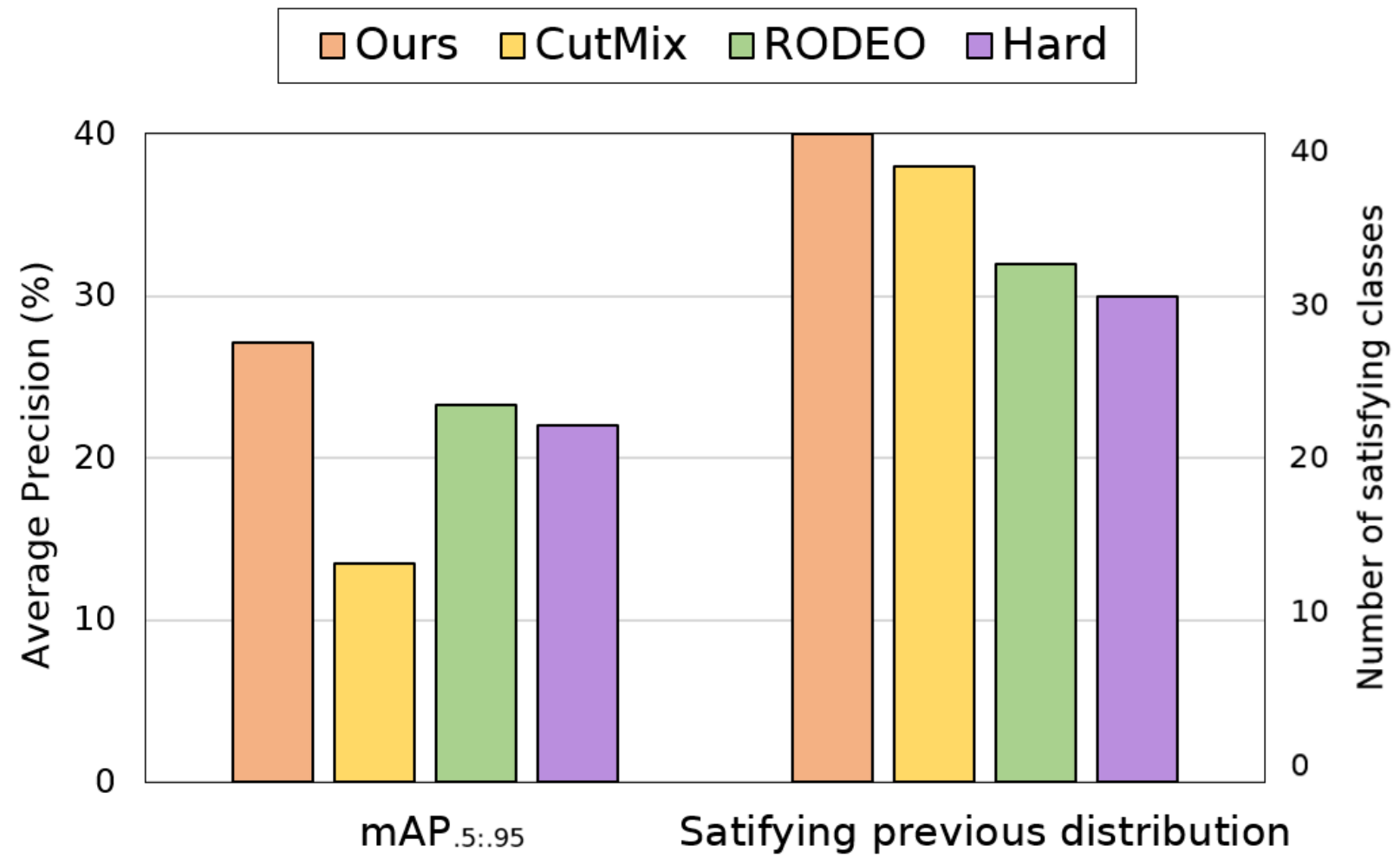}
}
\vspace{-1em}
    \caption{The final mean average precision ($mAP$. \%) and the number of classes that satisfy COCO's distribution at the 40+40 setup. We use the following formula $\left( \text{buffer capacity} \times  \frac{\text{Number of samples in } C_i}{\text{Total number of samples in } C_{1, \ldots, n}} \right)$ to check the distribution for the previous classes.
}
\vspace{-1em}
\label{fig:fig1}
\end{figure}

Most incremental methods~\cite{rebuffi2017icarl, guo2020randomsampling, kirkpatrick2017overcoming, rolnick2019experience, lopez2017gradient, koh2021blurry, bang2021rainbow} handle image classification task. We can also apply these incremental methods to object detection; however, due to varying labels for foreground objects in the scene, the strategies for object detection are relatively ineffective. 
Nevertheless, this task can play an important role in real-world applications. This allows us to adapt to environments where new object labels are constantly appearing. For example, when a new product is discovered, the detection system should recognize it while simultaneously detecting previous labels. Instead of completely retraining the model every time new labels appear, it helps to update the model to accommodate the unseen label incrementally. This greatly improves flexibility and persistence in real-world applications and saves computing resources. We call this work class incremental object detection (CIOD).

One of the most commonly used methods in CIOD is experience replay (ER)~\cite{chaudhry2018randomsampling, shieh2020CutMixReplay, acharya2020rodeo, liu2020multi, he2018exemplar, rolnick2019experience}. 
Random-based ER~\cite{chaudhry2018randomsampling, shieh2020CutMixReplay, guo2020randomsampling, rolnick2019experience} mitigates the complexity of multiple labels by simply randomly sampling from the previous data and building a buffer~\cite{rolnick2019experience} for integration with the new data. RODEO~\cite{acharya2020rodeo} and Hard~\cite{liu2020multi} suggested replay designed for CIOD, but it is still unclear whether they are the best strategy for preventing forgetting. Due to the lack of clarity on this effectiveness, we consider that there is room for enhancing these learning strategies, specifically \emph{replay strategy}.

In this paper, we propose an effective class-wise buffer training strategy, \sysname{}. Our methods consist of two buffer configuration components and a simple but efficient training approach. 
First, \emph{guarantee minimum} ensures the inclusion of a minimum quantity of each class sample, reflecting the class distribution of the prior dataset in the buffer. 
Second, \emph{hierarchical sampling} prioritizes samples with high number of unique labels and low loss when the buffer becomes full. This helps to retrain more diverse labels and optimize data to the trained model.
In terms of training approach, we propose \emph{circular experience replay} (CER) that deals with the asymmetry between current and prior tasks' data. It combines original ER training~\cite{rolnick2019experience, acharya2020rodeo, liu2020multi} and CER training, which are designed to avoid overfitting and to enhance prior knowledge. In Fig.~\ref{fig:fig1}, our method demonstrates the ability to accurately reflect the prior distribution in the buffer, as well as excellent performance. Our contributions can be summarized as follows:
\begin{enumerate}[label=\arabic*), itemjoin={\quad}]
    \item We introduce a buffer management strategy that is easily compatible with CIOD. The buffer manager operates the buffer based on two criteria: high number of unique labels and low loss, rather than any other single measures (e.g. many labels, randomness, etc.). We experimentally verified that it is the viable measure that reflects the tendency of the trained model.
    \item We propose an effective buffer training scheme, i.e. circular training, to overcome the imbalance caused by the limited capacity of the replay buffer and enhance previous detection performance. 
\end{enumerate}

\begin{figure*}[!t]
\centering{
\includegraphics[width=1\linewidth]{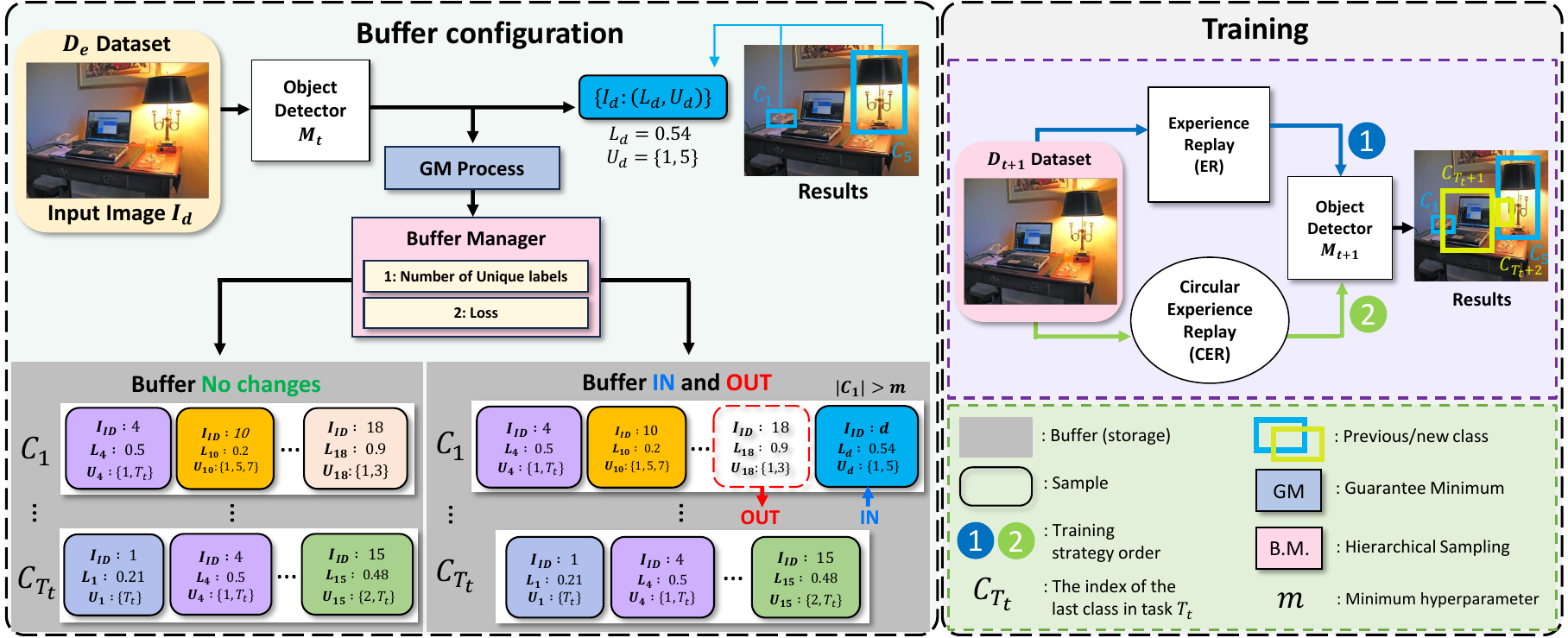}
}
    \vspace{-2em}
    \caption{The overall process. During the buffer configuration process, when the buffer is full, we perform the GM process to ensure coverage of all classes. Next, we select a candidate from the buffer to compare with new data based on two conditions. When training new data, we use the CER training strategy following ER training strategy within a specific epoch.}
    
\label{fig:model}
\end{figure*}

\section{Methods}
\label{sec:method}
\subsection{Overview}
\label{subsec:overview}
Our goal is to continually expand our knowledge by incorporating new labels while retaining previous knowledge in class incremental object detection (CIOD). The setting of CIOD consists of multiple tasks, each with a predefined number of object classes denoted as $T_t = {T_{1}, \ldots, T_{N}}$, where $N$ is the total number of tasks. 
Each task has its own corresponding dataset $\mathcal{D}_t$ which includes a set of input images $\mathbf{X}_t$ and corresponding labels $\mathbf{Y}_t$:
\vspace{-0.5em}
\begin{eqnarray}
    \tiny
   ~~\mathcal{D}_t\sim(\{\mathbf{X}_t:x_1, ..., x_{n_t} \in T_t\}, \{\mathbf{Y}_t:y_1, ..., y_{n_t}  \in T_t\})
\end{eqnarray}
where $n_t$ is the number of data contained in $T_t$, and $t$ is task index. Also, we use the buffer $\mathcal{B}$ which is a memory used for replay to store sample data (i.e. x, y). The $\mathcal{B}$ consists of data structured as follows:
\vspace{-0.2em}
\begin{eqnarray}
    \{I_{i}: (L_{i}, U_{i})\}
\label{eq:dataconfig}
\end{eqnarray}
where $I_{i}$ denotes the $i$-th image, $L_{i}$ is the associated sample loss, and $U_{i}$ signifies the list of unique classes in it.


Our effective buffer training strategy, \sysname{} has three main components: 1) guarantee minimum process to construct the representative image buffer (Sec.~\ref{subsec:GMprocess}), 2) hierarchical sampling for effective buffer configuration (Sec.~\ref{subsec:Hier}), 3) circular training for utilization of the buffer (Sec.~\ref{subsec:CER}). The overall flow is shown in Fig.~\ref{fig:model}. We will describe more details in the following subsection.

\begin{algorithm}[!t]
\SetAlgoLined
\setstretch{0.85}
\small 
\textbf{Input:} $K$, $m$, $\mathcal{D}_{t}$, $\mathcal{B}_{1:t-1}$, $\mathcal{M}_{1:t}$ \\
\textbf{define:} $\mathcal{B} \equiv \{I : (L, U)\}$ \hfill// \textcolor{blue}{buffer data format} \\
\textbf{define:} $\mathcal{D}_{e} \equiv \{x_1, ...,x_{N_e}\}$ \hfill// \textcolor{blue}{extra dataset format} \\
\vspace{0.4em}
$\mathcal{D}_{e} = \mathcal{B}_{1:t-1} \cup \mathcal{D}_t$ \textbf{if} $t > 1$ \textbf{else} $\mathcal{D}_t$ \hfill// \textcolor{blue}{extra dataset} \\

    \For{$d=1, \ldots, N_e$}{
        $I_{d}, U_{d} \leftarrow$ get\_info($d$)  \hfill // \textcolor{blue}{id, unique labels of $d$} \\
        $ L_{d} \leftarrow \mathcal{M}_{1:t}(d)$  \hfill // \textcolor{blue}{loss value of $d$} \\
        \eIf{$|\mathcal{B}| < K$}{
            $\mathcal{B} \leftarrow$ ($I_{d}$, $L_{d}$, $U_{d}$) \\
        }{  
            // \textcolor{blue}{pick all labels set. (e.g. $U_1, \ldots, U_K$) } \\
            $U_{B} \leftarrow$ get\_all\_unique\_labels\_set($\mathcal{B}$) \\
            \vspace{0.3em}
            // \textcolor{blue}{pick the labels that appear less than $m$ in $U_{d}$} \\
            $\mathcal{U} = \{u \in U_{d} \mid \text{count}(u, U_{B}) < m\}$ \hfill // \textcolor{blue}{Eq.~\ref{equation:u_c}} \\

            \eIf{$\mathcal{U} = \emptyset$}
                {
                    $\mathcal{R} \leftarrow \text{get\_samples}(\mathcal{B})$ \\} 
                {
                    $\mathcal{R} \leftarrow \text{get\_samples\_excluding\_labels}(\mathcal{B},  \mathcal{U})$ \\}
            $\mathcal{B} \leftarrow \text{\emph{BufferManager}}(\mathcal{B}, \mathcal{R}, \mathcal{U}, (I_{d}, L_{d}, U_{d}))$ \\        
            
}}

\textbf{Output: } $\mathcal{B}$ \\
\caption{\emph{Guarantee Minimum process}}
\label{Alg:DetailedGMMode}
\end{algorithm}
\subsection{Guarantee minimum process}
\label{subsec:GMprocess}
Data imbalance is a common problem in object detection tasks. Specifically, when creating a replay buffer, classes that are already under-represented in the data distribution may become scarcer, which can degrade detection performance. Therefore, it is important to ensure each class within the replay buffer has a minimum number of data. To address the issue, we propose the guarantee minimum (GM) method. This method maintains class-wise diversity in $\mathcal{B}$ and preserves the original data distribution $\mathcal{D}_{1:t-1}$ by ensuring a minimum of $m$ samples for every class.

\noindent\textbf{Data structure}. 
We generate an extra dataset $\mathcal{D}_{e}$ by combining $\mathcal{D}_{t}$ and $\mathcal{B}_{1:t-1}$, and uses input sample $d$ as Eq.~\ref{eq:dataconfig}. Given $I_d$ and $U_d$, we calculate the $L_{d}$ with pre-trained model $\mathcal{M}_{1:t}$. Since we employ a transformer-based detector, we construct the loss function as follows:
\vspace{-0.2em}
\begin{flalign}
    &L_{d}=L_\text{Bbox} + L_\text{GIoU} + L_\text{Label} 
\label{equation:loss}
\end{flalign}
where $L_\text{Bbox}$ and $L_\text{GIoU}$ represent L1 loss and generalized IOU loss~\cite{rezatofighi2019generalized} for bounding box. Additionally, $L_\text{Label}$~\cite{lin2017focal} is cross entropy with focal loss for label.
If the buffer $\mathcal{B}$ has not reached its maximum capacity $K$, the new data is directly added. However, once it attains full capacity, a strategic approach becomes necessary for data replacement.

\noindent\textbf{Guarantee process}. 
To replace the buffer samples with class-wise diversity, we first identify the sets of unique labels $U_B \sim \{U_1, \ldots, U_K\}$ in the $\mathcal{B}$. After that, we introduce the set $\mathcal{U}$ containing under-represented labels (i.e. class indexes) below a certain bound $m$:
\vspace{-1em}
\begin{flalign}
    &\mathcal{U} = \{u \in U_{d} \mid \text{count}(u, U_{B}) < m\}
\label{equation:u_c}
\end{flalign}
where, $u$ is an element of the unique labels from the input data $U_{d}$, and $m$ represents the minimum guarantee value. Then, we select the replacement candidates set $\mathcal{R}$ which contains samples without labels from $\mathcal{U}$ in $\mathcal{B}$. If $\mathcal{U}$ is empty (i.e. all classes above $m$), we choose all samples in $\mathcal{B}$ as replacement candidates $\mathcal{R}$. This approach ensures that our buffer reflects the overall class distribution, while also covering the rare labels more effectively. Finally, we use buffer manager employing hierarchical sampling (Sec.~\ref{subsec:Hier}) to compare $\mathcal{R}$ with a new sample. We summarize our GM algorithm in Alg.~\ref{Alg:DetailedGMMode}.



\begin{algorithm}[t]
\setstretch{0.85}
\SetAlgoLined
\small
\textbf{Input:} $\mathcal{B}$, $\mathcal{R}$, $\mathcal{U}$, $I_{d}, L_{d}, U_{d}$ \hfill // \textcolor{blue}{inputs from Alg.~\ref{Alg:DetailedGMMode}} \\
\textbf{define:} $\mathcal{B}, \mathcal{R} \equiv \{I : (L, U)\}$ \hfill // \textcolor{blue}{buffer \& candidates format} \\


    \vspace{0.4em}
    $\mathcal{R}_{\text{min\_U}} \leftarrow$ min\_U($\mathcal{R}$) \hfill // \textcolor{blue}{cond. 1: number of unique labels} \\
    
    $I_{\text{opt}}, L_{\text{opt}}, U_{\text{opt}} \leftarrow$ highest\_L($\mathcal{R}_{\text{min\_U}}$) \hfill// \textcolor{blue}{cond. 2: loss} \\
    \vspace{0.3em}
    \eIf{$\mathcal{U} = \emptyset$}{
        \eIf{$L_{\text{opt}} > L_{d}$}
            {
                del $\mathcal{B}[I_{\text{opt}}]$ \hfill // \textcolor{blue}{delete data in buffer} \\
                $\mathcal{B} \leftarrow$ ($I_{d}$, $L_{d}$, $U_{d}$) \hfill // \textcolor{blue}{insert new data to buffer} \\
            }
            {   no change \\}
    }
    {
        del $\mathcal{B}[I_{\text{opt}}]$ \hfill // \textcolor{blue}{delete data in buffer} \\
        $\mathcal{B} \leftarrow$ ($I_{n}$, $L_{d}$, $U_{n}$) \hfill // \textcolor{blue}{insert new data to buffer} \\
    }
    
\textbf{Output: } $\mathcal{B}$ \\
\caption{\emph{BufferManager}}
\label{Alg:BufferManager}
\end{algorithm}

\subsection{Hierarchical sampling strategy}
\vspace{-0.5em}
\label{subsec:Hier}
In this section, we introduce hierarchical sampling to create a buffer containing representative samples of the prior knowledge through two strategies: high number of unique labels and low loss. The high number of unique labels strategy~\cite{acharya2020rodeo} is used to diversify the buffer configuration, preserving previously learned labels within a limited capacity. However, when the buffer needs replacement, samples with an equally low number of unique labels are randomly replaced without specific conditions. Therefore, we use a low-loss approach for a more sophisticated configuration. In general, a low loss value indicate that the prediction is similar to the actual sample and the model has been well-trained on that particular sample. Thus, we prioritize data by using the loss for samples with the same number of unique labels. We utilize hierarchical sampling (summarized in Alg.~\ref{Alg:BufferManager}) to compare replacement candidates $\mathcal{R}$ and the input sample. We allocate an additional epoch to process all configuration procedures.

\begin{table*}[ht]
\centering
\caption{Incremental results for the COCO validation set using Deformable DETR in various scenarios. $T_1$ (40 or 70) represents the previous classes, and $T_{(1+2)}$ (80) denotes testing for all classes. The best result is highlighted in bold.}
\renewcommand{\arraystretch}{1.3}
\setstretch{0.8}
\resizebox{0.95\textwidth}{!}{
\begin{tabular}{c|l|cccccc|cccccc}
\hline
\multirow{2}{*}{Scenarios} &\multirow{2}{*}{Method}  & \multicolumn{6}{c|}{$T_{1}$(Old)} & \multicolumn{6}{c}{$T_{(1+2)}$(Overall)} \\ 
\cline{3-8} \cline{9-14}
 & & $mAP_{.5:.95}$ & $mAP_{.5}$ & $mAP_{.75}$ & $mAP_{S}$ & $mAP_{M}$ & $mAP_{L}$ & $mAP_{.5:.95}$ & $mAP_{.5}$ & $mAP_{.75}$ & $mAP_{S}$ & $mAP_{M}$ & $mAP_{L}$ \\ \hline
\multirow{5}{*}{\shortstack{70 + 10}} 
                & CutMix~\cite{shieh2020CutMixReplay} & 0.087 & 0.207 & 0.065 & 0.028 & 0.098 & 0.141 & 0.086 & 0.206 & 0.063 & 0.034 & 0.097 & 0.135  \\
                & RODEO~\cite{acharya2020rodeo}  & 0.064 & 0.109 & 0.066 & 0.042 & 0.097 & 0.091 & 0.094 & 0.151 & 0.100 & 0.056 & 0.127 & 0.137 \\ 
                & Hard~\cite{liu2020multi}  &  0.068 & 0.124 & 0.067 & 0.059 & 0.104 & 0.075 & 0.095 & 0.161 & 0.098 & 0.074 & 0.128 & 0.120 \\ 
                & Ours w/o CER & 0.179 & 0.288 & 0.192 & 0.089 & 0.209 & 0.238 & 0.190 & 0.304 & 0.203 & 0.097 & 0.218 & 0.261 \\ 
                & Ours  & \textbf{0.213} & \textbf{0.334} & \textbf{0.231} & \textbf{0.104} & \textbf{0.237} & \textbf{0.295} & \textbf{0.221} & \textbf{0.345} & \textbf{0.240} & \textbf{0.114} & \textbf{0.246} & \textbf{0.308}  \\ \hline\hline
\multirow{5}{*}{\shortstack{40 + 40}}
                & CutMix~\cite{shieh2020CutMixReplay} & 0.131 & 0.286 & 0.104 & 0.058 & 0.150 & 0.201 & 0.135 & 0.295 & 0.106 & 0.051 & 0.148 & 0.212  \\
                & RODEO~\cite{acharya2020rodeo} & 0.095 & 0.153 & 0.099 & 0.073 & 0.113 & 0.103 & 0.233 & 0.343 & 0.252 & 0.130 & 0.256 & 0.311\\ 
                & Hard~\cite{liu2020multi} & 0.072 & 0.131 & 0.072 & 0.070 & 0.107 & 0.059 & 0.220 & 0.332 & 0.239 & 0.121 & 0.250 & 0.285\\ 
                & Ours w/o CER & 0.168 & 0.271 & 0.176 & 0.099 & 0.199 & 0.194 & 0.270 & 0.405 & 0.293 & \textbf{0.144} & \textbf{0.297} & 0.367 \\ 
                & Ours & \textbf{0.222} & \textbf{0.356} & \textbf{0.234} & \textbf{0.125} & \textbf{0.255} & \textbf{0.296} & \textbf{0.271} & \textbf{0.419} & \textbf{0.294} & 0.136 & 0.296 & \textbf{0.376} \\ \hline\hline
\end{tabular}}
\label{table:result_all}
\end{table*}
\begin{figure}[t!]
\setstretch{1.2}
\centering{
\includegraphics[width=1\linewidth]{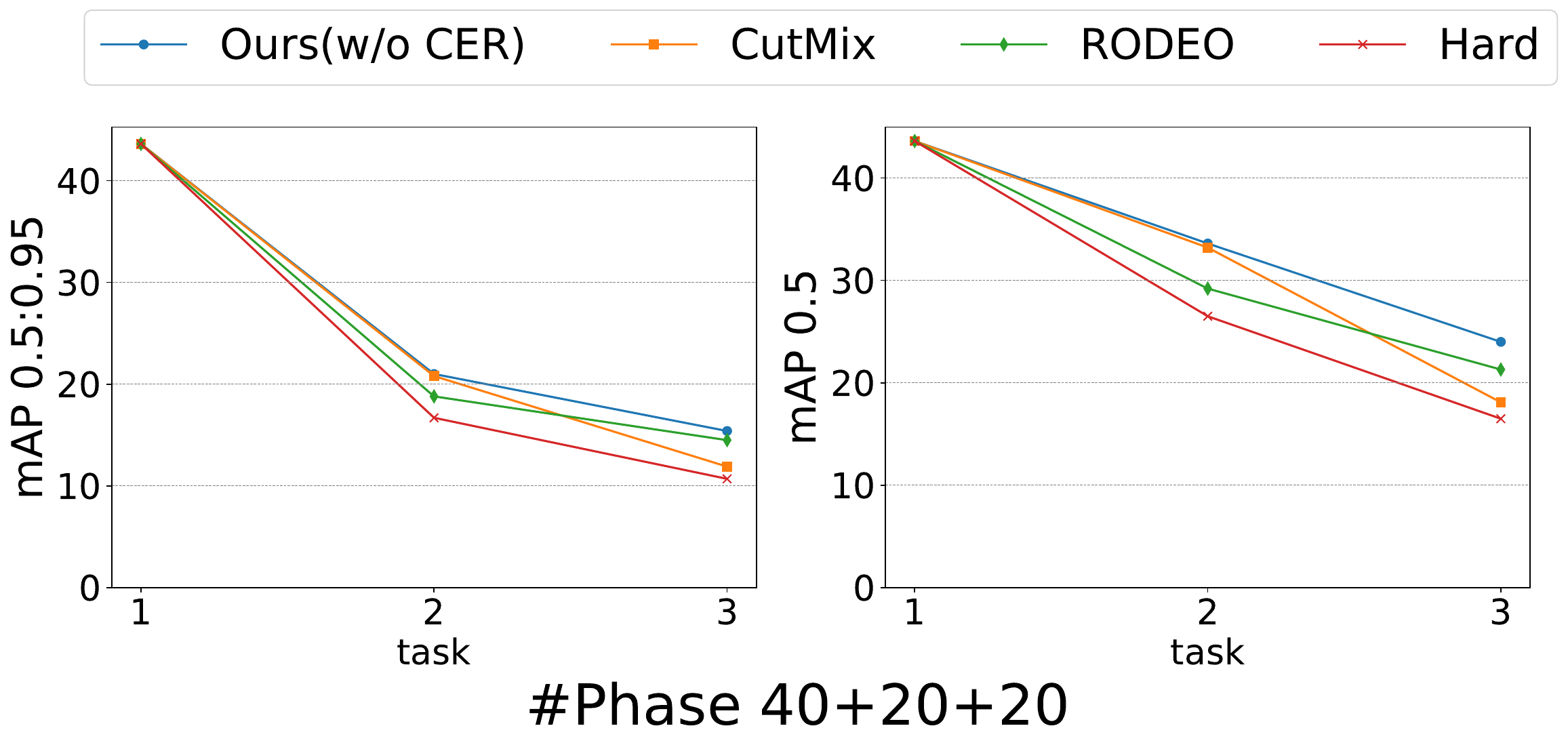}
}
\vspace{-2em}
    \caption{Multi-phase result.}
\label{fig:fig3}
\end{figure}

\subsection{Circular experience replay training}
\label{subsec:CER}
Previous CIOD replay methods~\cite{acharya2020rodeo, liu2020multi} used experience replay (ER) training method, which utilized a large buffer capacity to prevent forgetting a relatively small number of buffer data. However, this approach results in significant resource wastage. To address this issue, we propose the circular experience replay (CER) training strategy to make full use of the buffer which has limited capacity. First, we separates $\mathcal{D}_{t+1}$ and $\mathcal{B}_{1:t}$ to create the distinct training datasets. We then train the model with randomly selecting data from both datasets. The $\mathcal{B}_{1:t}$ is repeatedly utilized with uniform probability until all $\mathcal{D}_{t+1}$ is fully used. To enhance the utilization of previous information, we apply CER training following ER training.

\section{Experiments}
\label{sec:exp}
\subsection{Implementation and experiments}
\sysname{} is based on Deformable DETR~\cite{zhu2020deformable} trained from scratch on the COCO~\cite{lin2014microsoft} for 50 epochs in each task. All experiments are performed using 4 RTX3090 GPUs with batch size of 3. In the first, \emph{two-phase}, we incrementally train 40+40 and 70+10 divided classes. We evaluate the model on $T_{1+2}$ and $T_1$ to assess the degree of forgetting. In the second, \emph{multiple-phase}, we train 40+20+20 divided classes. Then, we test the model by combining the added classes. To ensure a fair comparison, we only extracted the replay components from various CIOD methods~\cite{shieh2020CutMixReplay, acharya2020rodeo, liu2020multi} that use replay and trained them using our baseline. We kept all conditions identical, except for the buffer composition (random~\cite{shieh2020CutMixReplay}, high number of unique labels~\cite{acharya2020rodeo}, many labels~\cite{liu2020multi} and training method (original  ER~\cite{acharya2020rodeo, liu2020multi}, CutMix~\cite{yun2019cutmix} based CutMix ER~\cite{shieh2020CutMixReplay}). In all our experiments, we set the buffer capacity at around 1\% (1200) of the COCO, and the least $m$ set at 1\% (12) of the buffer capacity.

\begin{table}[!t]
\centering
\caption{Comparison of the appropriate proportions of CER used with ER on COCO. The best result is highlighted in bold} 
\renewcommand{\arraystretch}{1.2}
{\begin{tiny}
\resizebox{0.45\textwidth}{!}{
\begin{tabular}{cc|c|c|c|c}
\hline 
\multicolumn{2}{c|}{phase} & \multicolumn{2}{c|}{7010} & \multicolumn{2}{c}{4040} \\
\hline
\multicolumn{2}{c|}{ER-CER Ratio} &  $T_1$ & $T_{(1+2)}$ &  $T_1$ & $T_{(1+2)}$ \\ \cline{1-6} 
ER &+ CER & AP & AP & AP & AP \\ \hline
40 & 10   & 0.168 & 0.183 & 0.172 & 0.253 \\
42 & 8    & 0.169 & 0.185 & 0.192 & 0.262 \\
44 & 6    & 0.188 & 0.199 & 0.210 & \textbf{0.271} \\
46 & 4    & 0.194 & 0.208 & 0.192 & 0.260 \\
48 & 2    & \textbf{0.213} & \textbf{0.221} & \textbf{0.222} & \textbf{0.271}  \\
\hline
\end{tabular}}
\end{tiny}}
\label{table:ERCER}
\end{table}

\subsection{Experimental results}
We analyze \emph{two-phase} results using the mAP metric on COCO dataset~\cite{lin2014microsoft}. In Table~\ref{table:result_all}, we qualitatively show that our approach \sysname{} achieved state-of-the-art results in the $T_{1}$ and $T_{1+2}$. 
Furthermore, our method (``ours w/o CER") performs well even without using the circular training strategy, in comparison to previous methods.
This indicates the effectiveness of our buffer configuration algorithm, which includes guarantee minimum processing and a hierarchical sampling strategy, in retaining previous knowledge $T_{1}$. As shown in Fig.~\ref{fig:fig3}, our method (``Ours w/o CER") also demonstrates good performance after training the last task in the \emph{multi-phase}. To ensure a fair, we exclusively employed ER training for all methods, excluding CER from our complete algorithm and omitting CutMix training~\cite{yun2019cutmix} used in CutMix~\cite{shieh2020CutMixReplay}. CutMix demonstrates comparable performance to our approach at task 2, but becomes less effective as the number of classes to be collected increases.

\subsection{Ablation}
\label{Ablation}
In Table~\ref{table:ERCER}, we demonstrate how the ratio of ER and CER is defined within the specified 50 epochs. The best performance is achieved with a 48:2 ratio in both the 70+10 and 40+40. Furthermore, Table~\ref{table:result_all} highlights that CER significantly improves performance in the 70+10 setup, where a larger number of classes need to be retained. The mAP increases from 0.190 to 0.221, compared to a smaller improvement from 0.270 to 0.271 in the 40+40.

\section{Conclusion}
\label{sec:conclusion}
In this paper, we propose an improved replay scheme to overcome the existing constraints in the class incremental object detection task. Our approach, \sysname{}, effectively manages the replay buffer with the guarantee minimum process and hierarchical sampling. In addition, we use a circular training strategy to address data imbalance. Our method demonstrates better performance in reducing catastrophic forgetting on the COCO dataset compared to existing methods. The ablation study demonstrates the optimal ratios for experience replay and circular experience replay. In future work, we aim to integrate our proposed method with other strategies to deal with the forgetting problem more effectively.

\newpage
\clearpage
{
    \small
    \bibliographystyle{ieeenat_fullname}
    \bibliography{main}
}


\end{document}